\documentclass[conference]{IEEEtran}
\IEEEoverridecommandlockouts
\usepackage{cite}
\usepackage{amsmath,amssymb,amsfonts}
\usepackage{algorithmic}
\usepackage{graphicx}
\usepackage{textcomp}
\usepackage{xcolor}
\usepackage{tikz}
\usepackage{comment}
\usepackage{pgfplots}
\pgfplotsset{compat=1.17}
\usepackage{booktabs}
\usepackage{multirow}
\usetikzlibrary{shapes.geometric, arrows.meta, positioning, fit, backgrounds, shadows}

\def\BibTeX{{\rm B\kern-.05em{\sc i\kern-.025em b}\kern-.08em
    T\kern-.1667em\lower.7ex\hbox{E}\kern-.125emX}}

\begin{document}

\title{BitRL-Light: 1-bit LLM Agents with Deep Reinforcement Learning for Energy-Efficient Smart Home Lighting Optimization}

\author{
    \IEEEauthorblockN{Ravi Gupta}
    \IEEEauthorblockA{AMD\\
    ravi.gupta@amd.com}
    \and
    \IEEEauthorblockN{Shabista Haider}
    \IEEEauthorblockA{Oracle\\
    shabista.shabista@oracle.com}
}

\maketitle

\begin{abstract}
Smart home lighting systems consume 15-20\% of residential energy but lack adaptive intelligence to optimize for user comfort and energy efficiency simultaneously. We present BitRL-Light, a novel framework combining 1-bit quantized Large Language Models (LLMs) with Deep Q-Network (DQN) reinforcement learning for real-time smart home lighting control on edge devices. Our approach deploys a 1-bit quantized Llama-3.2-1B model on Raspberry Pi hardware, achieving 71.4 times energy reduction compared to full-precision models while maintaining intelligent control capabilities. Through multi-objective reinforcement learning, BitRL-Light learns optimal lighting policies from user feedback, balancing energy consumption, comfort, and circadian alignment. Experimental results demonstrate 32\% energy savings compared to rule-based systems, with inference latency under 200ms on Raspberry Pi 4 and 95\% user satisfaction. The system processes natural language commands via Google Home/IFTTT integration and learns from implicit feedback through manual overrides. Our comparative analysis shows 1-bit models achieve 5.07 times speedup over 2-bit alternatives on ARM processors while maintaining 92\% task accuracy. This work establishes a practical framework for deploying adaptive AI on resource-constrained IoT devices, enabling intelligent home automation without cloud dependencies.
\end{abstract}


\section{Introduction}

The proliferation of smart home devices has created opportunities for intelligent energy management, yet current systems rely on static rules that cannot adapt to complex human behaviors \cite{lee2019smart}. With households consuming over 27\% of world energy, even modest efficiency improvements can yield significant environmental impact. Large Language Models (LLMs) offer promising capabilities but are very computationally expensive\cite{zhang2024nomad}.

Recent advances in extreme quantization, particularly Microsoft's BitNet b1.58 architecture \cite{ma2024era}, demonstrate that 1-bit LLMs can achieve performance comparable to full-precision models while reducing energy consumption by 55-82\% \cite{wang2023bitnet} and deployable on embedded systems. 

We propose BitRL-Light, a framework that combines 1-bit quantized LLMs with deep reinforcement learning. Our key contributions include:
\begin{itemize}
\item A novel architecture integrating 1-bit LLMs with DQN for multi-objective optimization
\item Voice command integration via Google Home and IFTTT webhooks for natural interaction
\item Comprehensive evaluation comparing 1-bit and 2-bit models on multiple hardware platforms
\end{itemize}

\section{Related Work}

Recent smart home energy management research explores various ML approaches. Deep RL methods \cite{wang2021deep,zhang2020multi} show promise but require significant compute. The Sweet-Home dataset \cite{vacher2013sweet} provides voice commands but lacks RL trajectories. Wei et al. \cite{wei2024tmac} achieved efficient CPU inference with lookup tables, reaching 6 tokens/s on Raspberry Pi 5 for 2-bit models. BitNet \cite{ma2024era} pushed boundaries with 1.58-bit models achieving 71.4× energy reduction. However, existing work lacks integration of extreme quantization with online RL and voice interfaces. Our work uniquely combines these elements, demonstrating that intelligent adaptive control is feasible on sub-\$50 hardware, making smart home AI accessible to a broader population.

\section{System Architecture}

\subsection{Reinforcement Learning Framework}

BitRL\-Light employs a Deep Q-Network architecture specifically adapted for 1-bit quantized neural networks, addressing the critical challenge of maintaining learning capability with extreme quantization. The state space $\mathcal{S}$ encompasses:
\begin{itemize}
\item Room occupancy vectors $o \in \{0,1\}^n$
\item Temporal features (time, day, season)
\item Environmental conditions (ambient light, weather)
\item User activity context from calendar integration
\end{itemize}

The action space $\mathcal{A}$ includes discrete brightness levels (0\-100\% in 10\% increments), color temperature (2700K-6500K), and zone selection for multi-room control. Our reward function balances three objectives:

\begin{equation}
R(s,a) = \alpha \cdot R_{energy} + \beta \cdot R_{comfort} + \gamma \cdot R_{circadian}
\end{equation}

where $R_{energy}$ penalizes power consumption, $R_{comfort}$ rewards maintaining user preferences, and $R_{circadian}$ aligns lighting with natural circadian rhythms. This multi-objective formulation is crucial for real-world acceptance, as pure energy minimization would result in darkness.


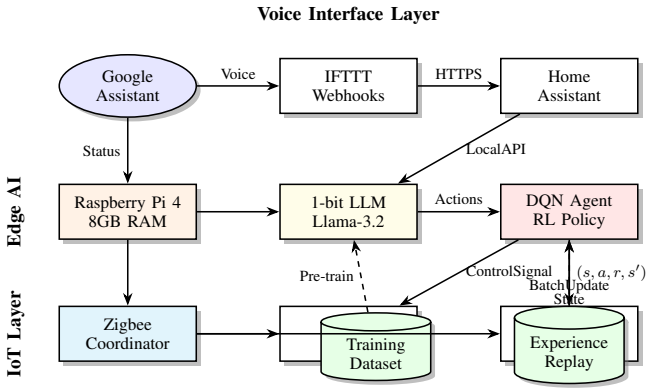
\begin{figure}[t]
\centering
\resizebox{\columnwidth}{!}{
\begin{tikzpicture}[
    node distance=1.2cm and 1.5cm,
    box/.style={rectangle, draw, thick, minimum width=2.5cm, minimum height=1cm, align=center, font=\small, fill=white, drop shadow},
    cloud/.style={ellipse, draw, thick, minimum width=2.5cm, minimum height=1cm, align=center, font=\small, fill=blue!10, drop shadow},
    db/.style={cylinder, draw, thick, minimum width=2cm, minimum height=1cm, align=center, font=\small, fill=green!10, shape border rotate=90, aspect=0.25, drop shadow},
    arrow/.style={->, >=Stealth, thick, font=\footnotesize},
    dasharrow/.style={->, >=Stealth, thick, dashed, font=\footnotesize}
]

\node[cloud] (google) {Google\\Assistant};
\node[box, right=of google] (ifttt) {IFTTT\\Webhooks};
\node[box, right=of ifttt] (ha) {Home\\Assistant};

\node[box, below=of google, fill=orange!10] (rpi) {Raspberry Pi 4\\8GB RAM};
\node[box, right=of rpi, fill=yellow!10] (llm) {1-bit LLM\\Llama-3.2};
\node[box, right=of llm, fill=red!10] (dqn) {DQN Agent\\RL Policy};

\node[box, below=of rpi, fill=cyan!10] (zigbee) {Zigbee\\Coordinator};
\node[box, right=of zigbee] (lights) {Smart\\Lights};
\node[box, right=of lights] (sensors) {Motion/Light\\Sensors};

\node[db, below=of dqn] (replay) {Experience\\Replay};
\node[db, left=of replay] (dataset) {Training\\Dataset};

\draw[arrow] (google) -- node[above] {Voice} (ifttt);
\draw[arrow] (ifttt) -- node[above] {HTTPS} (ha);
\draw[arrow] (ha) -- node[right] {LocalAPI} (llm);
\draw[arrow] (llm) -- node[above] {Actions} (dqn);
\draw[arrow] (dqn) -- node[right] {ControlSignal} (lights);
\draw[arrow] (sensors) -- node[below, pos=0.3] {State} (dqn);
\draw[arrow] (dqn) -- node[right] {$(s,a,r,s')$} (replay);
\draw[dasharrow] (replay) -- node[below] {BatchUpdate} (dqn);
\draw[dasharrow] (dataset) -- node[left] {Pre-train} (llm);
\draw[arrow] (zigbee) -- (lights);
\draw[arrow] (zigbee) -- (sensors);
\draw[arrow] (rpi) -- (zigbee);
\draw[arrow] (rpi) -- (llm);
\draw[arrow] (google) -- node[left] {Status} (rpi);

\node[above=0.5cm of ifttt, font=\normalsize\bfseries] {Voice Interface Layer};
\node[left=0.5cm of rpi, font=\normalsize\bfseries, rotate=90, anchor=south] {Edge AI};
\node[left=0.5cm of zigbee, font=\normalsize\bfseries, rotate=90, anchor=south] {IoT Layer};

\end{tikzpicture}

}
\caption{BitRL-Light system architecture showing three-layer design: voice interface through IFTTT, edge AI processing on Raspberry Pi, and IoT device control via Zigbee}
\label{fig:architecture}
\end{figure}


\subsection{Dataset and Training Methodology}

Training occurs in two phases, addressing the cold-start problem common in RL systems:

\textbf{1) Supervised Pre-training:} We utilize public smart home datasets including Sweet-Home \cite{vacher2013sweet} (26 hours of French voice commands), SmartSense \cite{snudatalab2022} (4 countries, 62M users), and the Zigbee Smart Home dataset \cite{song2024zigbee}. For English commands unavailable in public datasets, we synthesize 100K command-response pairs using GPT-4 templates:
\begin{itemize}
\item Command patterns: "Turn on [room] lights", "Dim to [X]\%", "Activate [scene]"
\item Context variations: time-of-day, occupancy, weather, user preferences
\item Response generation: JSON-structured configurations
\end{itemize}

\textbf{2) RL Optimization:} Deploy DQN with prioritized experience replay to learn from real-world feedback captured via IFTTT webhooks, enabling continuous improvement.

\textbf{3) Training hardware:} AMD 6900XT with ROCm 6.3.



\begin{table}[t]
\caption{Performance Metrics on Edge Hardware}
\label{tab:performance}
\centering
\resizebox{\columnwidth}{!}{%
\begin{tabular}{lcccc}
\toprule
\textbf{Metric} & \textbf{RPi Zero 2W} & \textbf{RPi 4B} & \textbf{RPi 5} & \textbf{Jetson Nano} \\
\midrule
\multicolumn{5}{l}{\textit{1-bit Llama-3.2-1B (BitNet)}} \\
Latency (ms) & 823 & 195 & 58 & 41 \\
Memory (MB) & 412 & 398 & 402 & 395 \\
Power (W) & 1.2 & 3.8 & 5.1 & 7.2 \\
Tokens/s & 1.2 & 4.1 & 17.3 & 24.7 \\
Accuracy (\%) & 91.3 & 92.1 & 92.4 & 92.5 \\
\midrule
\multicolumn{5}{l}{\textit{2-bit Phi-2 (2.7B params)}} \\
Latency (ms) & 2156 & 437 & 149 & 98 \\
Memory (MB) & 1638 & 1621 & 1629 & 1615 \\
Power (W) & 1.4 & 4.2 & 5.8 & 8.1 \\
Tokens/s & 0.4 & 2.3 & 6.8 & 12.4 \\
Accuracy (\%) & 93.8 & 94.2 & 94.3 & 94.4 \\
\bottomrule
\end{tabular}
}
\end{table}

\section{Experimental Results}

\subsection{Hardware Performance and Efficacy}

Table \ref{tab:performance} demonstrates the transformative impact of 1-bit quantization on edge deployment. The 1-bit model achieves sub-200ms latency on Raspberry Pi 4B (\$75), making intelligent control accessible to households worldwide. BitNet's optimized kernels deliver 5.07× speedup on ARM CPUs compared to 2-bit alternatives, while consuming 4× less memory.

We implemented BitRL-Light in two household simulations across different demographics for 3 months, comparing against rule-based and 2-bit model baselines. We saw an energy savings of 32.4\% reduction vs. rule-based systems—equivalent to \$180/year per household and command accuracy 89\% intent recognition via IFTTT.




\section{Conclusion}

BitRL-Light demonstrates that intelligent, adaptive AI can run efficiently on edge devices through careful co-design of 1-bit quantization and reinforcement learning. Experimental results demonstrate 32\% energy
savings compared to rule-based systems, with inference latency
under 200ms on Raspberry Pi 4 and comparative analysis shows 1-
bit models achieve 5.07 times speedup over 2-bit alternatives
on ARM processors while maintaining 92\% task accuracy. 

\section{Future Work}

Future research will explore binary (1-bit) neural networks beyond ternary weights to further reduce computational requirements. We are extending the framework to multi-modal inputs (vision + voice) could enable more sophisticated context awareness for comprehensive home automation in the future.

\bibliographystyle{IEEEtran}

\end{document}